# Novel Deep Learning Architecture for Predicting Heart Disease using CNN

Shadab Hussain
*Computer Science and Mathematics*
*Liverpool John Moores University, UK*
https://orcid.org/0000-0003-0521-3964

Santosh Kumar Nanda
Techversant AI-Data Science Team
Techversant Infotech Pvt Ltd
Trivandrum, Kerala, India
https://orcid.org/0000-0002-4112-5059

Susmith Barigidad
Computer Science and Engineering,
Santa Clara University, USA
sbarigidad@scu.edu

Shadab Akhtar
Computer Science and Engineering, GL
Bajaj Institute of Technology and
Management, India
shadabakhtar1993@gmail.com

Md Suaib
Computer Science and Engineering,
Saroj Institute of Technology and
Management, India
msuaib.569@gmail.com

Niranjan K. Ray
School of Computer Engineering KIIT
Deemed to be University Bhubaneswar-
24, Odisha
https://orcid.org/0000-0002-7197-4545

*Abstract:* **In the last few years, with increased population the most critical component of human life is healthcare. Compare to other deadly diseases, heart disease is one of the most lethal diseases, affecting the lives of millions of people worldwide. It is very important to detect heart disease must early so the loss of lives can be prevented. The availability of enormous amounts of data for medical diagnostics has aided in the development of complex learning-based models for automated early detection of cardiac problems. The classical machine learning approaches unable to generalize the new data sets which have not been seen in the training set. Therefore, the trained model has less accuracy in prediction stage. To minimize this issue, need to balance between training and testing datasets. This paper proposes a novel deep learning architecture using a 1D convolutional neural network for classification between healthy and non-healthy persons with balanced datasets to reduce the limitations of classical machine learning approach. Several clinical parameters are used for evaluating the risk contour in the patients which supports in early diagnosis. Various regularization methods are used to avoid overfitting in the proposed model. The proposed model achieves over 97% training accuracy and 96% test accuracy on the dataset. This is compared in detail with other machine learning algorithms using various performance parameters which proves the effectiveness of the proposed model.**

## I. Introduction

There has been considerable research in the field of healthcare in the last few years particularly after the Covid pandemic. According to the World Health Organization, heart disease is one of the worst diseases, accounting for the majority of human fatalities worldwide [1]. It is also observed that more than 24% of the deaths in India are due to various forms of heart disease [2]. As a result, there is a need to develop an early detection system that prevents mortality caused by cardiac disorders.

Heart disease, also known as cardiac disease, is caused by the constriction of the coronary arteries, which provide blood to the heart. There are methods like Angiography which is used for detecting heart diseases but it is very costly and is prone to certain reactions in a patient's body. This prevents the widespread use of these techniques in countries with large poor populations.

There is a need of developing healthcare products that provide quality results at an affordable rate. Healthcare organizations are also looking for clinical tests which can be performed without invasion at a cheap rate. The creation of a computer-based decision support system for the diagnosis of various diseases can assist organizations in meeting the needs of millions of people worldwide.

 The rapid growth of machine learning and deep learning algorithms has helped research in various industries including medical. The availability of large-scale medical diagnosis data has helped in training these algorithms. The clinical support system can be developed using these algorithms which helps in reducing cost and increasing accuracy [3].

Various clinical features can be utilized by machine learning algorithms for categorizing the risk profile of the patients. There are certain features like age, sex, heredity which are not in control while features like blood pressure, smoking, drinking habits are in control of the patient [2]. The proposed algorithm uses a combination of these features for categorizing healthy and non-healthy patients.

The rest of the paper is structured as follows: Section II discusses the available approaches of heart disease classification utilizing machine learning technologies. Section-III provides an explanation of the suggested architecture. Section-IV discusses the implementation specifics and findings.

## II. Literature Survey

A lot of work has gone into designing a heart disease diagnosis system for early identification using several clinical criteria. For identifying patients, many methods such as Logistic Regression, Support Vector Machine, Decision Tree, Random Forest, Artificial Neural Network, and others are utilized. This section summarizes those implementations.

S. Radhimeenakshi [4] proposed a Classification of cardiac disease using a decision tree and support vector machine. In terms of accuracy as measured by a confusion matrix, he concluded that the decision tree classifier outperforms SVM.

R.W.Jones et al [5] developed a strategy for predicting cardiac disease using an ANN. They used a self-applied questionnaire for training the neural network. The neural network contained three hidden layers and was trained using a backpropagation algorithm. The architecture was validated using the Dundee rank factor score and achieved a 98% relative operating characteristic value on the dataset.

Ankita Dewan et al. [6] compared the performance of genetic algorithms and backpropagation for training the neural network architecture. They concluded backpropagation algorithms perform better with a very minimum error on the dataset. SY Huang et al. [7] proposed a learning vector quantization algorithm for training the artificial neural network. They used 13 clinical features for training the network and achieved almost 80% accuracy on the dataset.

Jayshril S. Sonawane et al. [8] proposed a new artificial neural architecture that can be taught with a vector quantization technique and random order incremental training They also used 13 clinical features for training and attained an accuracy of 85.55 percent on the dataset. Majid Ghonji Feshki et al. [9] used four different classification algorithms for detecting cardiac diseases. They concluded that the PSO algorithm with neural networks achieved the best accuracy of around 91.94% on the dataset.

R. R. Manza et al. [10] proposed an ANN with a large number of neurons in the hidden layer which uses a Radial Basis Function. They obtained around 97% accuracy on this architecture. Saba Bashir et al. [2, 10] proposed a hybrid model for heart disease prediction which uses a combination of decision trees, SVM, and Naïve Bayes algorithms. They achieved 74% sensitivity, 82% accuracy, and 93% specificity.

P. Ramprakash et al. [1] proposed a deep neural network and $\chi^2$ statistical model for feature selection. They used various techniques to avoid overfitting and underfitting. They achieved 94% accuracy, 93% sensitivity, and 93% specificity. Turay Karayilan et al. [2] studied the performance of artificial neural networks with the various number of hidden layers. They achieved around 95.55% accuracy using five hidden layers.

It can be observed that most of the proposed systems use Artificial Neural Networks with some modifications. It is observed that these architectures are prone to overfitting so perform poorly on new data. So this paper proposes a new architecture using a one-dimensional convolutional neural network with dropout to avoid overfitting It uses the Cleveland database [12] which includes 13 characteristics for distinguishing between healthy and unhealthy individuals. The other classification algorithms are also implemented for verifying the performance of the proposed architecture using well-known performance measuring parameters.

A detailed explanation of the proposed architecture with the algorithms and techniques used in the next section.

III. PROPOSED ARCHITECTURE

This section describes the proposed architecture and all its constituent layers in detail along with the techniques used to optimize the architecture. It also gives some theoretical background about the 1-D convolutional neural network (CNN) which is central to the proposed architecture.

Conventional 2D CNN has become very popular in pattern recognition problems like Image classification and object detection [13]. CNNs are similar to ANN in which they consist of self-optimizing neurons which are trained to perform a certain task. This has led to the development of 1-D CNN which can operate on one-dimensional datasets or Time series data [13]. The proposed architecture using this concept of 1D CNN is shown in Fig. 1 below.

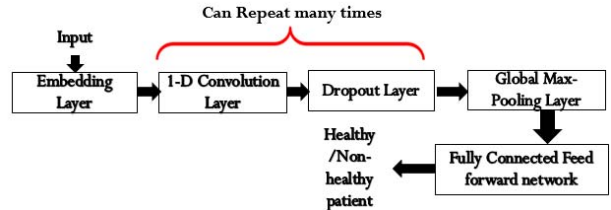

Fig 1: Proposed 1-D CNN Architecture

The input to the architecture will be the 13 characteristics that are crucial in the prediction of heart disease. These features are converted to a new representation called word embedding by the layer called Embedding Layer. It is similar to the Bag of Words concept used for Text data. It helps in a better representation of the dataset according to unique values present in each of the features. The Embedding layer's output is given to the 1D CNN layer for feature extraction.

1D CNN is very similar to conventional 2D CNN but the convolution operation is only applied to the one dimension which results in shallow architecture which can be easily trained on normal CPU or even embedded development boards [13]. The convolution operation helps in finding useful hierarchical features from the dataset which are useful in classification. The dimensions of the output features after 1D CNN can be calculated using the equation given below:

$$x = \frac{w+2p-f}{s} + 1 \qquad (1)$$

Where $x$ is the dimension of output features and $w$ is the size of input features. $f$ indicates the size of the filter used for convolutions. '$p$' indicates padding which are values added on the boundary before applying convolution. '$s$' indicates stride which is the value travelled after applying convolution operation.

The 1D convolution operation is a linear operation that is not useful in classifying nonlinear data. Most of the real-world dataset is nonlinear which requires some nonlinear operation after convolution. This nonlinear function is called an activation function. Some of the most common activation functions are the sigmoid, hyperbolic tangent, and rectified linear unit (RelU). The proposed architecture uses the RelU activation function which is easy to compute and allows faster computation. It also does not suffer from vanishing or exploding gradient problems.

There can be multiple convolution layers in the architecture followed by an activation function. The proposed architecture uses two 1-D convolution layers with 128 filters and filter sizes of 3. The output of the final convolution layer is passed through the global max-pooling layer which pools the maximum value from all the channels and reduces the dimension of output. The output of pooling is given to the fully connected layer with 256 neurons which extracts the useful features for classification. This layer is similar to the hidden layer is ANN. The final layer contains a single neuron which gives the classification probability. The final layer uses the sigmoid activation function as it directly gives the probability for binary classification.

The layer-wise details along with output feature dimensions and trainable weights of every layer are shown in Table 1.

Table 1: Layerwise CNN Architecture

| Type of the Layer | Output Dimension | No. of Weights and biases |
|---|---|---|
| Embedding Layer-1 | (13, 300) | 45600 |
| Dropout Layer-1 | (13, 300) | 0 |
| 1D Convolution Layer-1 | (13, 64) | 57664 |
| Dropout Layer-2 | (13, 64) | 0 |
| 1D Convolution Layer-2 | (13, 64) | 12352 |
| GlobalMaxPooling Layer-1 | (None, 64) | 0 |
| Dense Layer-1 | (None, 256) | 16640 |
| Dense Layer-1 | (None, 1) | 257 |
| Number of parameters: 132,513 | | |
| Number of Trainable parameters: 132,513 | | |
| Number of Non-trainable parameters: 0 | | |

The proposed 1D CNN architecture contains around 0.13 million trainable parameters which will get adapted during the training of the network. It was observed that general CNN architecture overfitted the training data meaning that training accuracy was very high and validation accuracy was low. The dropout technique was introduced to remove overfitting. It removes random neurons with a certain probability during training which allows the different networks to be trained at every iteration. This will help in the network not being too dependent on any single neuron of the network. The dropout layer has been introduced after each trainable layer in the proposed architecture. The addition of the dropout layer helped the training and test accuracy to be very similar which points to the network adapting well to data that it has not seen.

The next section describes the implementation details and results obtained after training the proposed architecture.

## IV. IMPLEMENTATION AND RESULTS

The suggested architecture for heart disease prediction was built with the scikit-learn and Keras libraries, which enables the implementation of various machine learning and deep learning algorithms. The development setup includes an Intel i5 CPU and 8GB of RAM. It also contains a GeForce 940 GPU, which aids in the training of the architecture. The paper uses the Cleveland database [12] which has 303 samples of patients with 14 different features. The dataset is split into two halves where 80% is used for training and the remaining 20% is used for validation. The clinical parameters (features) used for classification in the dataset are explained in Table 2.

Table 2: Dataset details

| Sr No. | Clinical Parameters | Range of Values |
|---|---|---|
| 1 | Age of Patient | 29-77 |
| 2 | Gender of a patient | 0 = Female<br>1 = Male |
| 3 | Category of pain in Chest | 0 = Atypical Angina<br>1 = typical Angina<br>2 = Asymptotic<br>3 = Non Angina |
| 4 | Thallium Scan | 3=normal<br>6=fixed<br>7=reversible effect |
| 5 | BP of a Patient | 94-200 |
| 6 | Exercise related to rest which causes ST depression | 0 – 6.2 |
| 7 | Cholesterol Level | 126-564 |
| 8 | Blood Suger in fasting condition | 0 if < 120<br>1 if >= 120 |
| 9 | ECG report in resting condition | 0 = Normal<br>1 = ST-T wave abnormalities<br>2 = left ventricular Hypertrophy |
| 10 | Pulse rate | 71-202 |
| 11 | Angina caused by exercise | 0 = No<br>1 = Yes |
| 12 | Number of major vessels which are colored by Fluoroscopy | 0-3 |
| 13 | The gradient of ST segment in peak exercise condition | 0= un sloping<br>1=flat<br>2=down sloping |
| 14 | Heart Disease | 0 = No<br>1 = Yes |

Some of the attributes have missing values for some of the examples. . For training our architecture, the missing values have been substituted with the attribute's average value. Most of the traditional classification architectures require all the attributes in the same range. This dataset has attributes in different ranges so a standardization technique is applied which converts all the attributes into the same range. It subtracts all the attribute values with the average value of the attribute and divides them by the standard deviation of the attribute. The final attribute is the true label for the patient whether he/she has heart disease or not. The dataset is a little unbalanced in the sense that there are more negative examples compared to positive as shown in Fig. 2 below.

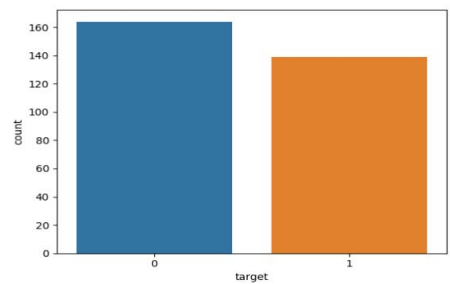

Fig. 2: Distribution of labels in the dataset

As can be seen in Fig. 2, there are 163 negative examples and 140 positive examples in the dataset. The first 13 attributes are used as input features for classification. The proposed architecture is trained for 150 epochs. The batch size is taken as 32. The binary cross-entropy is calculated between the true value and the predicted value for calculating loss for optimization. This function has to be minimized using some optimization algorithm to achieve convergence. The Adam optimization algorithm is used for training as it provides faster convergence and does not zigzag around the local minima [14].

ADAM uses exponentially weighted gradients as well as exponentially weighted square gradients for updating the weights at each iteration.

Exponentially decaying averages of past gradients is calculated by:

$$\mu_t = \alpha_1 \mu_{t-1} + (1 - \alpha_1) G_t \quad (2)$$

Where $\mu_t$ is the momentum term at timestamp 't', $\alpha_1$ is constant which is taken as 0.9 and $G_t$ is the gradient at timestamp 't'. Exponentially decaying averages of past squared gradients is calculated by:

$$V_t = \alpha_2 V_{t-1} + (1 - \alpha_2) G_t^2 \quad (3)$$

Where $V_t$ is the velocity term at timestamp 't', $\alpha_2$ is constant which is taken as 0.99 and $G_t$ is the gradient at timestamp 't'.

Bias correction $\hat{\mu}_t = \frac{\mu_t}{1-\alpha_1^t}$ and $\hat{V}_t = \frac{V_t}{1-\alpha_2^t}$ and then, update parameters using Adam's update rule:

$$W_{t+1} = W_t - \frac{\eta}{\sqrt{\hat{V}_t}+\epsilon} \hat{\mu}_t \quad (4)$$

Where ε is constant with a very small value which avoids division by zero and $W_t$ is a parameter value at timestamp 't'.

The training and test accuracy after each epoch is shown in Fig. 3 below:

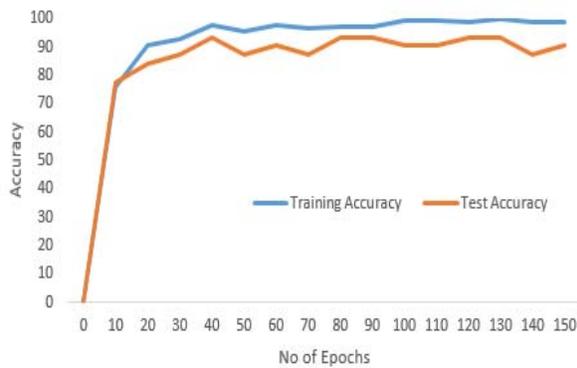

Fig. 3: Accuracy after each epoch using Adam Optimizer without Dropout

The proposed architecture achieves training accuracy of 98.9% and test accuracy of 90.32%. There is a large gap between training and test accuracies which indicates overfitting in the architecture. Dropout layers are added after every trainable layer in the architecture with a probability of neurons being removed is 0.3. The train and test accuracy at every epoch for the modified architecture is shown in Fig. 4.

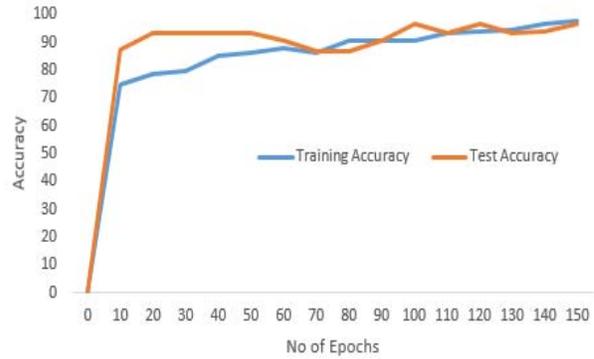

Fig. 4: Accuracy after each epoch using Adam Optimizer

The proposed architecture with dropout achieves training accuracy of 97.79% and test accuracy of 96.77%. Some other well-known classification algorithms are also implemented for comparing the performance of the proposed architecture. The detailed comparison table is shown in Table 3 below:

Table 3: Performance Comparison of Proposed Architecture with other Classifiers

| Algorithm | Training Accuracy | Test Accuracy | Precision | Recall | F1 Score | AUC |
|---|---|---|---|---|---|---|
| Logistic Regression | 86.36 | 80.32 | 85 | 65 | 73.5 | 78.0 |
| Naïve Bayes | 86.77 | 78.68 | 78.26 | 69.23 | 73.46 | 77.47 |
| SVM | 92.56 | 80.32 | 85 | 65.3 | 73.9 | 78.4 |
| Decision Tree | 100 | 77.04 | 73.07 | 73.07 | 73.07 | 76.53 |
| Random Forest | 99.17 | 77.04 | 77.23 | 65.38 | 70.83 | 75.54 |
| LightGBM | 99.58 | 77.04 | 83.33 | 57.69 | 68.18 | 74.56 |
| XGBoost | 100 | 78.68 | 84.21 | 61.53 | 71.11 | 76.48 |
| Artificial Neural Network | 88 | 78.68 | 78.26 | 69.23 | 73.46 | 77.47 |
| **Proposed Architecture (1D CNN)** | **97.79** | **96.77** | **94.73** | **100** | **97.29** | **96.15** |

The accuracy is calculated as the ratio of the total number of correctly predicted examples to total examples. It can be from the table that the proposed architecture is the best performing architecture in terms of test accuracy. Some other architectures perform well on the Training set but perform very poorly on the test set. The accuracy can alternatively be represented as a confusion matrix. It has a number of correctly classified examples in the diagonal and wrongly classified examples elsewhere. The confusion matrix for the given architecture is shown below in Table 4.

Table 4: Confusion Matrix

|  |  | Predicted Class | |
|---|---|---|---|
|  |  | 0 | 1 |
| True Class | 0 | 24 | 2 |
|  | 1 | 0 | 36 |

When the dataset is unbalanced then sometimes accuracy does not give a correct idea about the performance measure of the architecture. So, the performance is also measured in terms of other performance measuring parameters like F1 Score, precision, recall and Area under the receiver operating characteristic curve.

Precision is an indication of how many positive predictions are correct whereas recall identifies how many actual positive examples are correctly identified. There is always a tradeoff between precision and recall so a new performance measuring parameter F1 score is introduced. F1 score measures the harmonic mean of these two which gives a balance value between precision and recall.

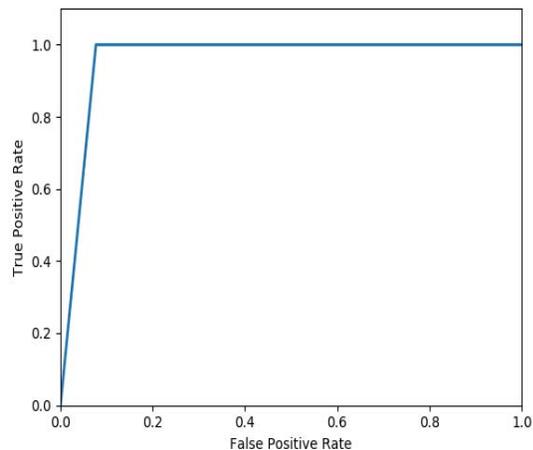

The last parameter AUC is a measure of the area under the ROC curve. The ROC curve is shown in Fig. 5.

Fig. 5: ROC curve

The ROC curve is a graph that indicated the relationship between the false positive rate and the true positive rate. The proposed architecture has a ROC curve that is very near to the ideal curve which indicates the good performance of the architecture on the test set.

The proposed architecture performs well in terms of all of these performance parameters. The proposed architecture is also verified on new data which is not available on either train or validation set. It achieves good performance on new data as well. The statistical importance of each feature in classification is also observed.

## V. CONCLUSION

The purpose of this paper is to use a computer-assisted technique to detect cardiac problems early. A 1D convolutional neural network design for predicting heart disease is proposed in this paper. It also contains an Embedding layer which converts the feature vector into new vector embedding which helps in classification. The proposed architecture is implemented as a software system on a computer that can help in the early diagnosis of cardiac disease at a cheap cost and with high accuracy. The architecture uses overfitting avoidance techniques which help the performance of unseen data. The performance of 1D CNN architecture is best among all other classification algorithms like Logistic Regression, Naïve Bays, SVM, Decision Tree, Random Forest, LightGBM, XGBoost, and ANN. More and more parameters can be included in the system which can help in classifying heart disease more accurately. It can also be integrated with wearable sensor readings for real-time prediction of heart diseases.